# Sushi Dish : Object detection and classification from real images

Yeongjin Oh[1]
Seoul National University
oyj9109@snu.ac.kr

Seunghyun Son[1]
Seoul National University
sonsong2001@snu.ac.kr

Gyumin Sim[1]
Seoul National University
sim0629@snu.ac.kr

## Abstract

*In conveyor belt sushi restaurants, billing is a burdened job because one has to manually count the number of dishes and identify the color of them to calculate the price. In a busy situation, there can be a mistake that customers are overcharged or undercharged. To deal with this problem, we developed a method that automatically identifies the color of dishes and calculate the total price using real images. Our method consists of ellipse fitting and convolutional neural network. It achieves ellipse detection precision 85% and recall 96% and classification accuracy 92%.*

## 1. Introduction

Conveyor belt sushi restaurant is popular format of sushi restaurants. People eat sushi freely from conveyor belt and pay for what they ate when they leave the restaurant. Problem occurs when people try to calculate the total price that they have to pay. Normally a color of dishes indicates the price of the sushi and employee in the restaurant has to count the number of each color of dishes to charge customers. However, this process is time consuming and stressful job. Also this process can give customer impression of unreliability. This is human oriented job and there can be a mistake which is related to money. In this paper, we propose a system that can automatically detect dish number and color from the real images taken by cell phone cameras.

Our proposed system consists of two major parts; detector and classifier. We find dish objects from the real image by ellipse detection because dishes in the image can be expressed as ellipses ignoring perspective distortion. After the detection, we apply convolutional neural network based classifier to the dish objects to find out what colors of the dishes are, which is related to the price of sushi. Our ellipse detector is capable of detecting dish and approximating the missing dish and classifier is robust to different illumination condition, shadows, stains such as sushi sauce in the dish, and even errors from the ellipse detector.

To get the data set, we collected data from restaurant, Niwa, and took more than 200 pictures of piled dishes with various combinations of different colors, illumination conditions, and perspectives. Those images were used for tuning the detector and training the classifier. After implementing the system, authors tested system with 15 images.

## 2. Detection

Our dish detector consists of the three stages: edge processing, ellipse detection, and reconstruction. In edge processing, we get smooth curvature which is strongly possible to be portion of ellipses. With the smooth curvature, we fit many ellipses and filter out outliers of them. From the detected ellipses, we reconstruct missing dishes by predicting the parameters. The details of each stage will be explained in the following subsections.

### 2.1 Edge Processing

After extracting an edge map from the real image, our goal is to process the edge map so that it can be used directly for ellipse detection. Using the information of the connected edge pixels or edge contours improves the possibilities of generating stronger cues for ellipses and for this, **Prasad [1]** defines 'smooth curvature' which is possible portion of ellipse. We follow 4 steps to extract smooth curvatures from the edge map.

#### 2.1.1 Edge detection

For the edge detection, we pre-process the input images. The images are converted to gray scale and resized not to be larger than 800 pixels length of bigger side because too large images may have too many edges even in the area with texture. Then, we perform histogram equalization to the resized images in order to correctly detect the edges even if they were taken on the dim illumination condition. After this, Canny edge detector **[2]** is applied to the pre-processed image with the parameters: low hysteresis threshold $T_L = 0.2$, high hysteresis threshold $T_H = 0.2$, and standard deviation for the Gaussian filter $\sigma = 1$.

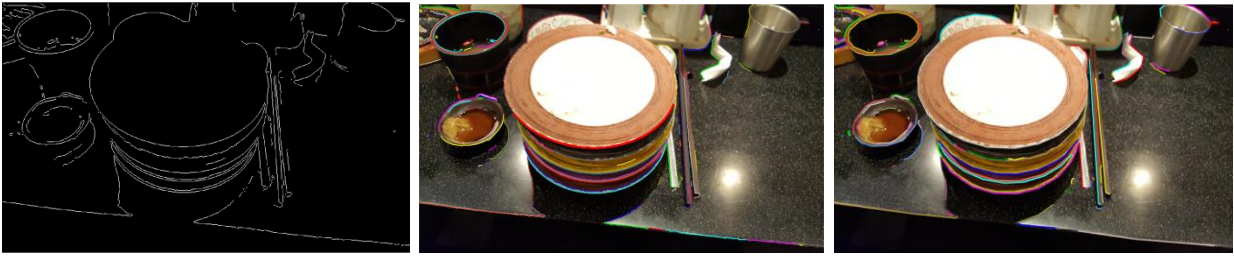

(a) Input edge map without branch     (b) The image with edge contours     (c) The image with smooth curvatures

Figure 1. The result of edge map processing, edge contour extraction, and smooth curvature division after line segment fitting

### 2.1.2 Edge contour extraction

To use the information of the connected edge contours, we have to extract continuous, non-branching edge contours. Thus, we apply 'thin' morphological operator to the edge map and remove all junctions to make sure of getting branchless edges. We also remove all isolated pixels so that we can assume two types of edge: an edge with distinct start and end point, a loop.

For an edge with distinct end points, we can easily find one of end points by following connected points. We mark all trace of points from the start point and follow connected point until getting stuck at the other end point. This work gives us an edge contour which consists of continuous points. After extracting all edges with distinct end points, there only remain loops. By selecting an arbitrary point as an end point among unmarked points in the edge map, a loop can be also extracted as an edge contour with two end points using the same work. Figure 1 shows the result image with edge contours.

### 2.1.3 Line segment fitting

Each edge consists of many continuous points and it makes the remaining processes take much more time. Since a few points more than five are enough to represent an ellipse, we reduce the number of points which represent an edge contour. By representing the edge contours using piece-wise linear segments, we can make it without loss of accuracy for ellipse detection. We use the **Ramer–Douglas–Peucker algorithm [3]** to approximate a curve into a set of line segments.

Let us consider an edge contour $e=\{P_1, P_2, ..., P_n\}$, where $e$ is an edge contour with end points $P_1$ and $P_n$. The line passing through a pair of end points $P_1(x_1, y_1)$ and $P_n(x_n, y_n)$ is given by :

$$x(y_1 - y_n) + y(x_n - x_1) + y_n x_1 - y_1 x_n = 0 \quad (1)$$

Then the deviation $d_i$ of a point $P_i(x_i, y_i)$ from the line passing through the pair is given as:

$$d_i = |x_i(y_1 - y_n) + y_i(x_n - x_1) + y_n x_1 - y_1 x_n| \quad (2)$$

Using the equation (2), we find a maximal deviation point $P_{MAX}$ and split an edge at that point. We repeat these steps until maximum deviation becomes small enough. Choosing 2 pixels as a deviation threshold gives us appropriate number of points representing line segments. Figure 2 shows the result of line segment fitting.

### 2.1.4 Smooth curvature division

Since the curvature of any elliptic shape changes continuously and smoothly, we intend to obtain edges with smooth curvature. The term smooth curvature is defined as a portion of an edge which does not have a sudden change in curvature, either in terms of amount of change or the direction of change **[1]**.

By splitting edge contours at either of sharp-turn point and inflexion point, we can get smooth curvatures. We choose $90°$ as a sharp-turn threshold for the amount of change. The inflexion point is defined as a point at which the direction changes.

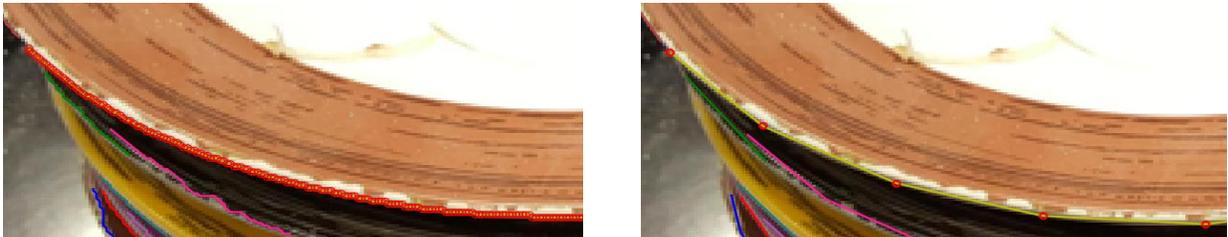

(a) Edge contours with continuous points     (b) Edge contours with line segments

Figure 2. Edge contours represented by continuous points, and line segments. Red circles show how many points are on an edge.

## 2.2 Ellipse Detection

We assume that the input images have little perspective distortion so that the border of dishes can be counted as ellipses. An ellipse has five parameters: the coordinate of center $(p, q)$, the major radius $A$, the minor radius B, and the orientation α; it can be represented by the equation (3):

$$\begin{bmatrix} x \\ y \end{bmatrix} = \begin{bmatrix} p \\ q \end{bmatrix} + \begin{bmatrix} \cos(\alpha) & -\sin(\alpha) \\ \sin(\alpha) & \cos(\alpha) \end{bmatrix} \begin{bmatrix} A\cos(\theta) \\ B\sin(\theta) \end{bmatrix} \quad (3)$$

We implement the ellipse fitting by using the code of **Richard [4].** It fits ellipses by solving least squares for the result segment of the edge processing, which is smooth curvature with more than 5 points.

However, the result of ellipse fitting has many false positives, which are not part of dishes but detected as ellipses. As shown in the Figure 3, there are many obstacles on the table and they are also detected as ellipses.

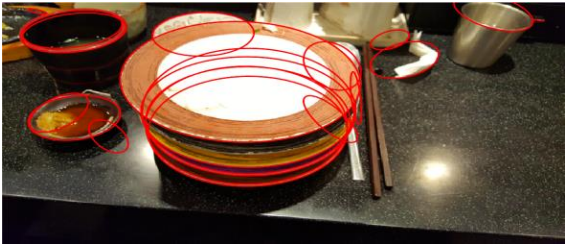

Figure 3. The result of ellipse fitting applied to smooth curvatures in Figure 1-(c). It has many false positives.

In order to filter them out, we make an assumption that input images show just one tower of stacked dishes and they are stacked vertically. Then, the correct ellipses for dishes have similar parameters of $p$. Also, they should have similar parameter values for $A$ and α because the all dishes have the same size and the same orientation. We adopt the idea of RANSAC to filter out the outliers for each of parameters $p$, $A$, and α. As a result, we get a clear tower of stacked ellipses.

Another problem is that the border of the dishes has a little thick white area. So, fitted ellipses are doubly detected for most of the dishes. We set a proper threshold as the minimal gap between a pair of the $y$-coordinates of the bottom-most point of each ellipse, and use it to identify whether the ellipses are doubly detected or not. After sorting the ellipses in descending order of $y$-coordinate of the bottom-most point, if two consecutive ellipses were not apart from each other less than the threshold, one of them would be thrown out; we choose the lower one. The result of elimination of false-positives is shown in Figure 4.

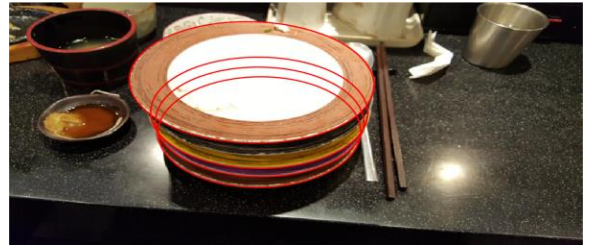

Figure 4. The result of RANSAC and double line elimination.

## 2.3 Reconstruction

Some dishes have many short fitting segments instead of a long one and it gives so small evidence that the dishes are not detected. To reconstruct the missing dishes, we predict an ellipse for each dish and find evidences to get an optimal fitting ellipse by calculating error between the ellipse and fitting segments. For this work, the assumption that dishes are stacked in a single tower is necessary.

### 2.3.1 Prediction

To predict missing dishes, we use ellipse parameter matrix which consists of parameters of all ellipses:

$$E = \begin{bmatrix} p_1 & q_1 & A_1 & B_1 & \alpha_1 \\ \vdots & \vdots & \vdots & \vdots & \vdots \\ p_n & q_n & A_n & B_n & \alpha_n \end{bmatrix} \quad (4)$$

Using parameters $[p_i\ q_i\ A_i\ B_i\ \alpha_i]$ representing an ellipse $E_i$, we find the bottom-most point $y_i$ according to $y$-coordinate on the ellipse $E_i$. Under the assumption that $y_i's$ are close to the linear sequence with small distortion of perspective, we can find index of missing dish by computing the gap between each pair of $y_i$ and $y_{i+1}$.

After finding index of missing dish, we predict an ellipse using the tendency of each parameter for the bottom-most and the intermediate dishes. Since the top-most dish has strong edge, it is rarely not detected. Then, we find fitting segments around each prediction. Figure 5 shows prediction ellipses and fitting segments.

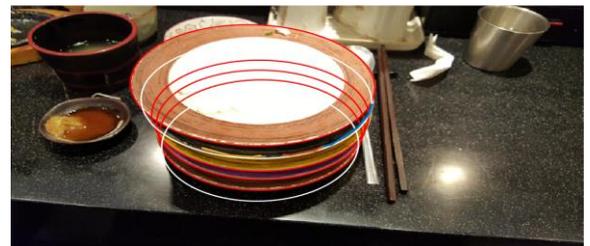

Figure 5. Prediction ellipses and fitting segments. White ellipses represent prediction using the ellipse parameter matrix of red ellipses.

### 2.3.2 Reconstruction

Before reconstructing the missing dish, we find an optimal ellipse fitting the dish and check whether it has strong evidences to avoid making false positives.

Given a prediction ellipse and fitting segments, we find an optimal ellipse which has the smallest error from fitting segments varying the prediction parameters $p_i, q_i, A_i, B_i$. Then, to reconstruct the optimal ellipse, we check some conditions:

1. Fitting segments covers more than 10% of perimeter of the optimal ellipse.
2. Error between fitting segments and the optimal ellipse is small enough; we choose 0.1 as the threshold value.
3. All points in fitting segments has the error smaller than 0.2.

If the optimal ellipse meets all three conditions, we regard it has strong evidence and reconstruct it. The result of reconstruction is shown in Figure 6.

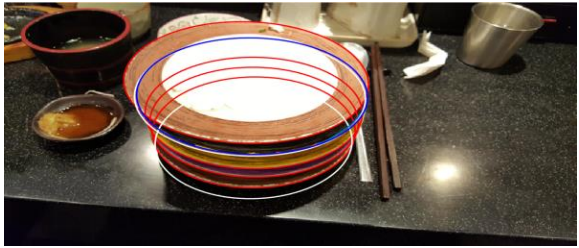

Figure 6. The result of reconstruction. A blue ellipse represents the optimal ellipse with strong evidence.

### 2.4 Evaluation

In the tuning data set, only 88 images meet our assumption and there are 461 dishes. Before the reconstruction stage, 361 dishes are correctly detected, and there were 6 false positives which are not actual dishes but detected as dishes. After the reconstruction 391 dishes are correctly detected, and there were 14 false positives. The precision and recall are summarized in Figure 7. Although the reconstruction slightly increases the false positive ratio by 1.82%p, it significantly increases the precision by 6.51%p. For the testing data set, with the reconstruction, the precision was 84.52% and the recall was 95.95% which are similar to the result for the tuning data set (84.82% and 96.54%).

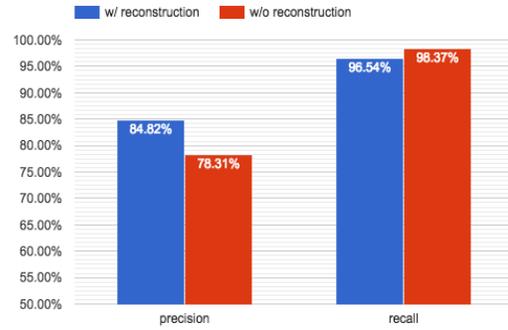

Figure 7. The precision and recall for the tuning data set with/without reconstruction.

### 3. Classification

For dish color classification, we use multi-layer convolutional neural network **[5], [6]** which is variant of neural network **[7]** which works well for image classification tasks. There are other methods for color detection. **Baek et al. [8]** proposed using 2D histogram from HSV color space and SVM. **Rachmadi et al [9]** proposed deep learning approach for vehicle color classification and showed that CNN works well on color classification. We use end-to-end network that automatically find useful features. This approach can deal with real images that suffer from different illumination condition, shadows, polluted dish image and even error from ellipse detector. Classification pipeline consists of 4 parts. Input definition, data augmentation, classifier architecture, and evaluation.

### 3.1 Input Image

Input to classifier is generated from results of ellipse detector. After ellipse detection, we transform ellipses to a circle by homography. For each dish, transformed circle is subtracted by upper transformed circle. The results are shown in Figure 8. Each image's dimension at this point is [100x100x3]. After this process, each image is cut in half and only the lower half of the image is fed into the network. This is done for the reduction of dimensionality and variance over images. Lower half of the image has all the information for classification and dish that was stacked at the top layer only has the upper part. After this process, image's dimension is [50x100x3] and all images are manually labeled with 8 classes of colors and set information.

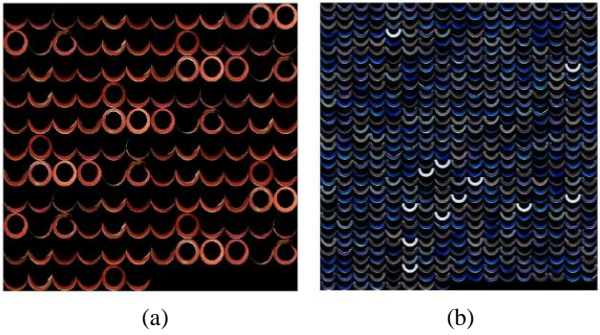

(a)                   (b)

Figure 8. The result of feature extraction.
(a) Features with dimension [100x100x3]
(b) Features after subtraction with dimensions of [50x100x3]
Original color of dishes is blue but it looks white or even black because of different illumination conditions and shadows.

### 3.2 Data Augmentation

To train deep network, there has to be sufficient number of training data so we synthesize training images. And to train network that is robust to noises, we add some noise to the images. Collected labeled data wasn't sufficient for deep network and number of images for each class was unbalanced. We use stochastic gradient descent method for updating weights. But difference of the number of images between classes was too high. For example, the number of brown colored dish images was 4 times larger than the number of red one. To make probability of each class to be in the batch even, we duplicate data to balance number of images among 8 classes. After this process, there were 1,031 images for training and validation.

To double existing training data, left and right of images are flipped to generate new data. After that, gaussian noise is applied with zero mean and 0.001 as variance. Applying gaussian noise is shown in Figure 9. After this process, we had 4,124 images for training and validation.

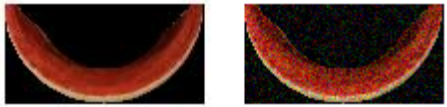

Figure 9. Left is original image and right is image with gaussian noise applied.

### 3.3 Classifier Architecture

We use multi-layer convolutional neural network which consist of 3 convolution layers and one fully connected layer with softmax. For pooling, 2x2 max pooling with stride 2 is used and for activation function, rectified linear unit is used. Whole model is shown in Figure 10.

First convolution layer takes [50x100x3] input. But we don't use augmented data directly. Data is subtracted by the data mean. Special thing about first layer is that it uses skewed filter to produce squared output. Followed by max pooling layer.

Second convolution layer is also followed by max pooling layer. Second layer's weight is [5x5x20x50] with stride 1.

Third convolution layer's weights are [4x4x50x500] with stride 1. Followed by ReLU layer [10]. After this layer, we get 500-dimensional feature vector that is fed into fully connected layer.

Last layer is fully connected layer with softmax. It classifies 500-dimensional feature vector into 8 classes. 8 classes used for classification is shown in Figure 11.

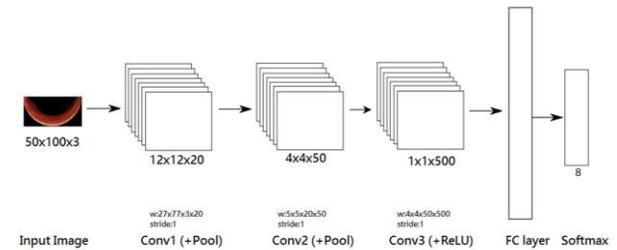

Figure 10. CNN classifier architecture. 500-dimensional feature vector is fed into fully connected layer.

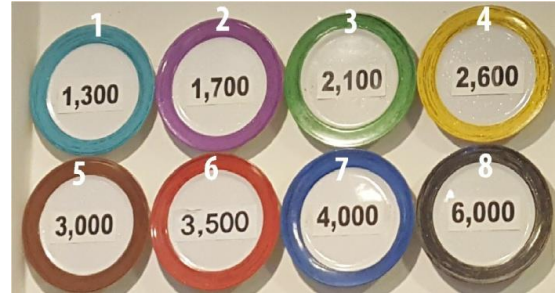

Figure 11. The labels of 8 classes used for training.

### 3.4 Evaluation

We trained our model with stochastic gradient descent method. Testing our model with 15 novel dish images, we got test accuracy 91.55%, 65 correct predictions of 71 dishes. Total confusion matrix for test image is shown in Figure 12. Training time and loss graph is shown in Figure 13. We found that our model has many points of improvement because this result is obtained from only 447 training images before data augmentation. It is very small compared to popular datasets that have more than ten thousand raw training images.

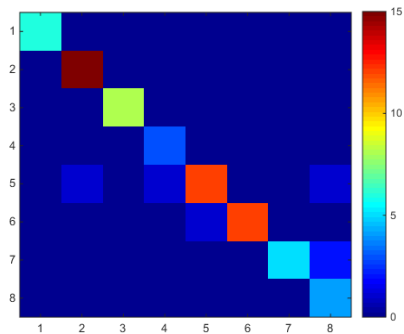

Figure 12. Confusion matrix for test image. Color close to red means more instance.

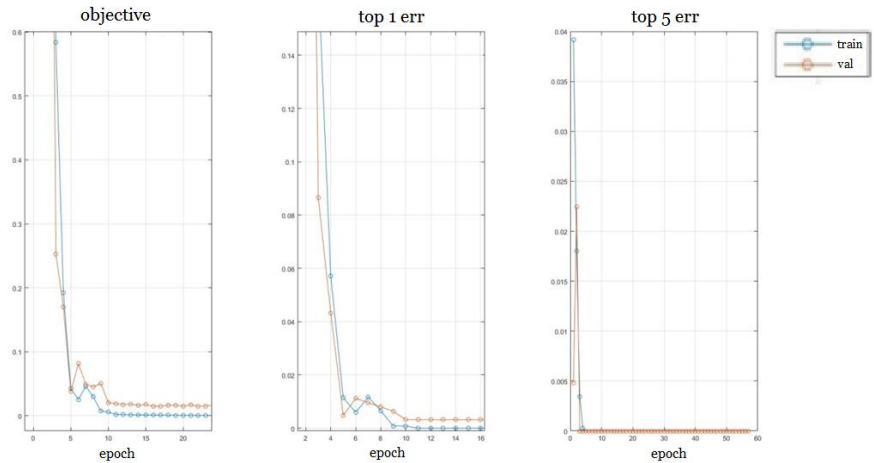

Figure 13. Training time- loss graph. validation error and train error sharply decreases.

4. Conclusion

In this paper, we present the sushi dish recognition system using ellipse detection and convolutional neural network. Our dish detector achieves precision 85% and recall 96% and classifier achieves accuracy 92%. This performance is considered meaningful because this result is from very small dataset. There were only 447 instances of dish images which are relatively small compared to popular datasets. With a slight improvement of current performance, our system could work on the field because real prototype is likely to have verification method from the user such as asking users (both employees and customers) about the suspicious images that have probability of being other classes in the softmax layer. The real working system may be able to be improved by itself by applying reinforcement learning on the classification. Also, applying some kind of machine learning approach to the detector would be another interesting study and a good attempt to improve the detection precision on the fly without further manual tuning by human.

Appendix

The full source code of our implementation can be found on GitHub: https://github.com/YeongjinOh/Sushi-Dish